\documentclass[letterpaper, 10 pt, conference]{ieeeconf}

\IEEEoverridecommandlockouts
\usepackage{cite}

\usepackage{amsmath, amsthm, amssymb,amsfonts}
\usepackage{textcomp}
\usepackage{stmaryrd}
\usepackage{upgreek}
\usepackage{bm}
\usepackage[linesnumbered,ruled,vlined]{algorithm2e}
\theoremstyle{definition}
\newtheorem{definition}{Definition}
\theoremstyle{remark}

\usepackage[pagewise]{lineno}
\usepackage{multirow}
\usepackage{subcaption}
\usepackage{comment}
\usepackage{graphicx}
\graphicspath{{./figures/}}
\usepackage{subcaption}

\usepackage{xcolor}
\newcommand{\remove}[1]{}

\title{\LARGE \bf
Causality-Based Parametric Control Barrier Function \\for Safe Multi-Vehicle Interaction
}

\author{Yiwei Lyu$^{1*}$, Caleb Chang$^{2*}$ and John M. Dolan$^{3}$%
\thanks{* These authors contributed equally.}
\thanks{$^{1}$The author is with the Department of Computer Science and Engineering, Texas A$\&$M University, College Station, TX, 77845 USA. Email: {\tt\small yiweilyu@tamu.edu}}%
\thanks{$^{2}$The author is a visiting researcher from the School of Electrical and Computer Engineering, Georgia Institute of Technology, Atlanta, GA, 30332 USA. Email: {\tt\small calebwychang@gatech.edu}}%
\thanks{$^{3}$The author is with the Robotics Institute, Carnegie Mellon University, Pittsburgh, PA, 15213 USA. Email: {\tt\small jdolan@andrew.cmu.edu}}%
}

\begin{document}

\maketitle
\thispagestyle{empty}
\pagestyle{empty}

\begin{abstract}
Safe control has been widely studied in various safety-critical applications, for instance, autonomous driving. In order to ensure the autonomous vehicle does not collide with other vehicles, it is essential to obtain an accurate expectation of surrounding vehicles' behavior and react adaptively. Instead of assuming fully cooperative and homogeneous vehicles using the same safety-critical controllers, recent works have been exploring different data-driven approaches to model the neighboring vehicles' underlying controllers with observed data. However, existing works either suffer from 1) the inter-vehicle influence during the multi-vehicle interaction, which makes it hard to determine the causality of surrounding vehicles' behavior in controller modeling, or 2) being dominated by the worst-case analysis, which may lead to overly conservative behavior. In this paper, we extend the prior work on Parametric-Control Barrier Function (Parametric-CBF) to multi-robot interactions with embedded causality inference to explicitly reason over the inter-vehicle influence. Given the learned Causality-based Parametric-CBF, we present an adaptive safety-critical controller that allows the ego vehicle to safely react to surrounding vehicles with the learned expectation. We demonstrate that by leveraging the motion flexibility among multi-vehicle systems, task efficiency can be greatly improved in various interaction-intensive scenarios.

\end{abstract}\vspace{-.5cm}

\section{Introduction}
In light of the need to operate in interactive environments with other scenario participants, e.g., other robots or humans, robots must ensure the safety of autonomy by correctly understanding others' capabilities and planning their own strategies in a reactive manner. To achieve these goals, robots must generate accurate predictions of other agents and reasonable reactions in response. In the domain of autonomous driving, how to anticipate other vehicles' behavior and leverage the anticipation to design controllers for ego vehicles while preserving formal safety guarantees has been a critical challenge.

There has been significant progress in trajectory prediction for road participants, primarily through data-driven approaches that predict the future movements of other vehicles~{\cite{mozaffari_deep_2022},\cite{grover2022semantically}}. However, these predictive methods, which rely on current and past states, often fail to accurately capture how other robots will dynamically adapt to various conditions, such as the presence of obstacles, environmental changes, and interactions with other agents. This limitation is particularly problematic in safety-critical scenarios, where understanding the interactive capabilities and possible reactions of other vehicles is crucial.

\begin{figure}
     \centering
    \includegraphics[width=.5\textwidth]{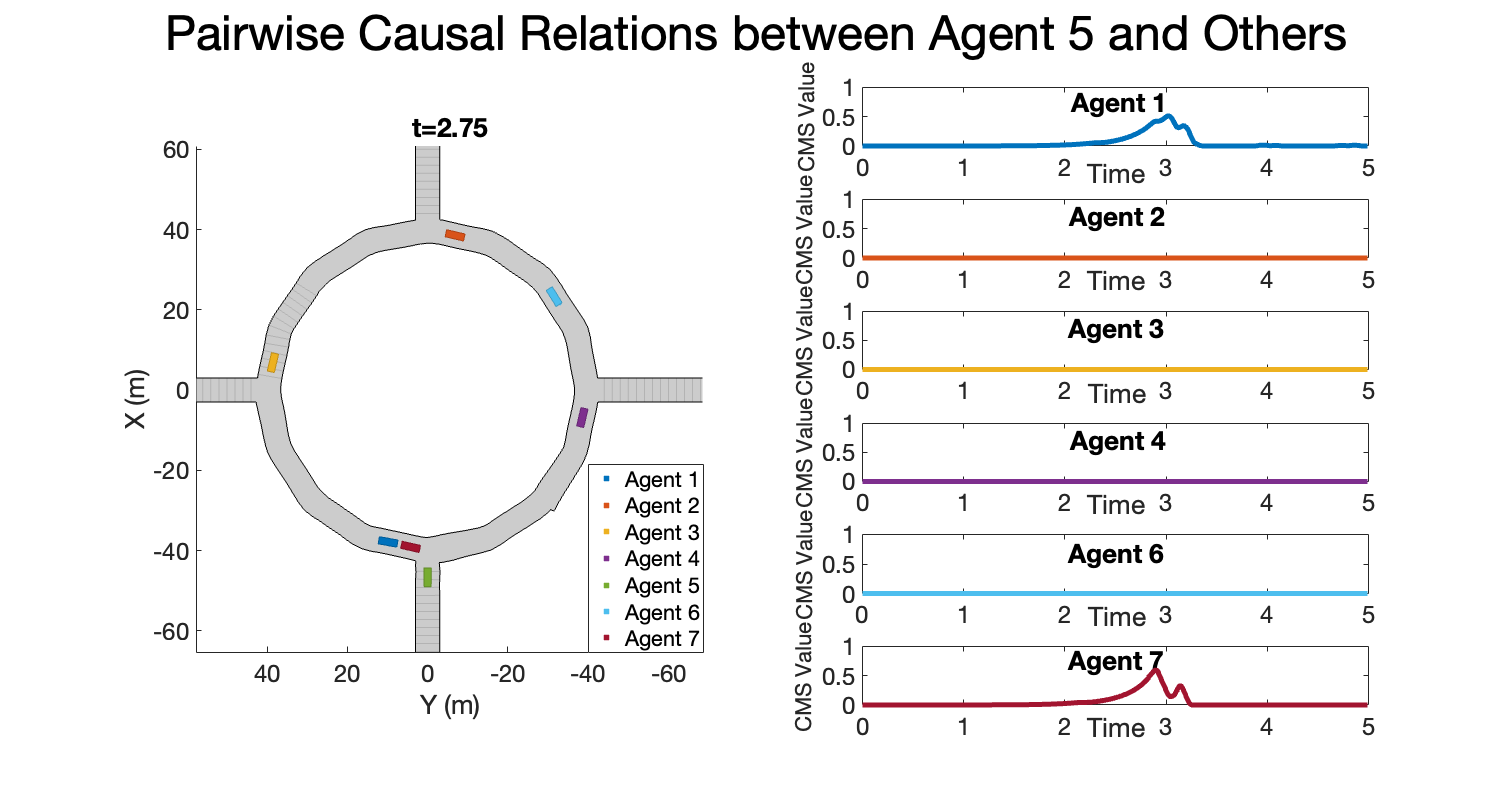}
     \caption{\footnotesize Cross Map Smoothness (CMS)-based causality inference for target vehicle Agent 5 in a 7-vehicle roundabout scenario. Left: A snapshot of the interaction scenario. Right: The CMS scores for each pairwise interaction involving the target vehicle.}
     \label{fig:CMS}
 \end{figure} 
To address these challenges, learning the underlying controllers of other vehicles has emerged as a more effective approach~\cite{qin_learning_2021,yang_enhancing_2024}. By gaining insights into the adaptive behaviors of robots in response to environmental changes, this class of methods enables more robust and flexible planning and decision-making. Furthermore, it allows us to explicitly model different driver characteristics underlying these behaviors, which can effectively facilitate proactive and expressive interactions when we design safe controllers for ego vehicles. Several attempts have been made to model the underlying controllers of other vehicles to better represent and interpret various driving styles~\cite{saveriano2019learning,robey_learning_2020}. 

Despite recent advances, existing approaches face several fundamental limitations. A recurring challenge is the trade-off between computational efficiency and behavioral expressivity: models rich enough to capture diverse driving styles are often too expensive for real-time deployment, whereas simpler formulations cannot represent important behavioral nuances. Many methods, such as~\cite{lyu_adaptive_2022}, attempt to learn the underlying controller and driver characteristics from direct observations of two-car interactions. These approaches typically rely on the assumption that all demonstrations are generated under saturated constraints, so that the observed actions directly reflect the enforcement of safety requirements rather than nominal control objectives. 

While~\cite{cosner2023learning} relaxes this assumption by incorporating likelihood maximization of driver characteristics in expert demonstrations, it still presumes that the data come from either a single expert or a set of experts with consistent styles. In practice, however, driving data are often heterogeneous~\cite{yarlagadda2022heterogeneity}; without a mechanism to evaluate which portions of the demonstrations are actually relevant, the resulting inaccuracies in learning can lead to unsafe or overly conservative behavior. 

Moreover, much of this prior work has been limited to two-vehicle scenarios, leaving open questions about scalability to realistic, multi-agent environments. In such settings, the causal dependencies~\cite{granger_investigating_1969,shojaie_granger_2022 } among interacting vehicles introduce an additional layer of complexity, making it difficult to determine when and why safety constraints are activated and which demonstrations are most informative for learning. Addressing these challenges is critical for developing models that can accurately and efficiently capture the interactive dynamics of heterogeneous multi-vehicle systems.

In this paper, our \textbf{main contributions} are, \textbf{1)} we introduce a causality detection algorithm to identify pair-wise interactions between vehicles on the road, enabling the robot to automatically reason over which portion of demonstrations should be used for learning without assumptions on constraint saturation and homogeneous experts, \textbf{2)} we extend prior work on Parameteric-CBF \cite{lyu_adaptive_2022} with embedded causality reasoning to enable learning of driving characteristics in a data-driven approach under motion uncertainty, and \textbf{3)} we present a comprehensive safety-critical control framework with our learning-enabled Parametric-CBF, which encourages courteous and efficient driving behavior while limiting perturbations to surrounding vehicles to improve overall task performance and collective safety. \vspace{-.3cm}

\section{Preliminaries and Notations}
\subsection{Background on CBFs and Parametric-CBFs}
Control Barrier Functions (CBFs) \cite{ames2019control} are used to guarantee safety in nonlinear systems (such as an affine nonlinear system) by the forward invariance of the safety set. The general nonlinear affine system is given by:
\begin{equation} \footnotesize
    \Dot{x}=f(x)+g(x)u + \epsilon, \quad \epsilon \sim \mathcal{N}(0,\sigma)
\end{equation}
where $x\in\mathcal{X}\subset\mathbb{R}^n$ and $u\in\mathcal{U}\subset\mathbb{R}^m$ are the system state and control input with f and g assumed to be locally Lipschitz continuous. We introduce a Gaussian disturbance $\epsilon$ to account for uncertain motions in the form of aleatoric uncertainty \cite{hullermeier_aleatoric_2021}, \cite{noauthor_aleatoric_nodate}.
A desired safe set $x\in\mathcal{H}$ can be denoted by the Control Barrier Function $h(x)$:
\begin{equation} \footnotesize
    \mathcal{H}=\{x\in\mathbb{R}^n:h(x)\geq0\}
\end{equation}
where $h(x)<0$ indicates a safety violation.
A formal definition  is presented below:

\begin{definition}
    (Control Barrier Function) Given a dynamical
    system (1) and the set $\mathcal{H}$ defined in (2) with a continuously differentiable function $h : \mathbb{R}^n\rightarrow \mathbb{R}$, then $h$ is a control barrier function (CBF) if there exists a class $\mathcal{K}$ function for
    all $x \in \mathcal{X}$ such that $\underset{u\in\mathcal{U}}{sup}\{\Dot{h}(x,u)\}\geq-\kappa(h(x))$,
    where $\Dot{h}(x,u)=L_fh(x)+L_gh(x)u$ with $L_f$ and $L_g$ as the Lie derivatives of $h$ along the vector fields $f$ and $g$ in the nonlinear affine system.
\end{definition}
 A commonly selected class $\mathcal{K}$ function is $\kappa(h(x)) = \gamma h(x)$~\cite{ames2019control,lyu2021probabilistic}, where $\gamma \in \mathbb{R}_{\geq 0}$ is a design parameter that regulates system behavior near the boundary $h(x) = 0$. This leads to the admissible control set: $\mathcal{B}(x) = \left\{ u \in \mathcal{U} : \dot{h}(x, u) + \gamma h(x) \geq 0 \right\}.$
As shown in~\cite{ames2019control}, any control input $u \in \mathcal{B}(x)$ renders the safe set $\mathcal{H}$ forward invariant, i.e., if $x(0) \in \mathcal{H}$, then $x(t) \in \mathcal{H}$ for all $t > 0$. However, the simple linear form $\kappa(h(x)) = \gamma h(x)$ can be limiting when capturing complex system behavior near the boundary.  Parametric Control Barrier Function (PCBF) framework~\cite{lyu_adaptive_2022} generalizes this formulation by effectively increasing the expressiveness of both $\gamma$ and $h(x)$, enabling more nuanced safety constraints and adaptive system responses.

\begin{definition}
    (Parametric-Control Barrier Function) Given a dynamical system (1) and the set $\mathcal{H}$ defined in (2) with a continuously differentiable function $h : \mathbb{R}^n \rightarrow \mathbb{R}$, then $h$ is a Parametric-Control Barrier Function (Parametric-CBF) for all $x \in \mathcal{X}$ such that $\underset{u\in\mathcal{U}}{sup}\{\Dot{h}(x,u)\}+\alpha H\geq0$,
    where parameter vector $\alpha=[\alpha_1, \alpha_2, ..., \alpha_q]\in\mathbb{R}^q$ with $\forall \alpha_p\in\mathbb{R}^{\geq0}$ for $p\in[q]$. $H(x)=[h(x), h^3(x), ..., h^{2q-1}(x)]^T, q\in\mathcal{N}$.
\end{definition}
As proved in~\cite{lyu_adaptive_2022}, the Parametric CBF retains all the theoretical guarantees of the vanilla CBF. $\alpha$ denotes the {safety behavior vector}, as it characterizes how rapidly the system approaches the boundary $\partial \mathcal{B}$. In the context of driving, $\alpha$ reflects the aggressiveness of a driver's behavior. In Section~III.C, we demonstrate how $\alpha$ can be learned from data through a regression-based approach.

\vspace{-.3cm}
\subsection{Background on Causality}
The study of causality seeks to explain the cause-and-effect relationships between two events. It identifies the cause-and-effect pair of events among many events and determines how strong the influence of the causal event is on the affected event. In this paper, we use the specific notion of Granger Causality (G-Causality) that associates predictability with causality in a time series setting \cite{shojaie_granger_2022}. Granger defines causality as follows \cite{granger_investigating_1969}:

\begin{definition}
    \label{def:causality}
    (Causality) Given a stationary stochastic system $A^t$ and $Hist(A^t)$, which is the set of past values $\{A^t-j, j=0,1,2,\dots, \infty \}$, $P^t(A|B)$ is the optimal predictor of $A^t$ using the set of values $B^t$. The predictive error is $\epsilon^t(A|B)$ and its variance is denoted by $\sigma^2(A|B)$. Let $U^t$ be all the information in the universe since time $t-1$ and $U^t-Y^t$ denote all the information apart from the time series $Y^t$. If $\sigma^2(X|Hist(U))<\sigma^2(X|Hist(U-Y))$, we say that $Y$ is causing X, denoted by $Y^t\rightarrow X^t$. $Y^t$ is causing $X^t$ if we can predict $X^t$ using all available information $U$. Conversely, even if we can predict $X^t$ using all available information $U$ without $Y^t$, then a causal relationship does not stand.
\end{definition}
Intuitively, two state variables are said to have a causal relationship if the state of one variable influences the dynamics of the other. In the case of $Y \rightarrow X$, the dynamics of $X$ are sensitive to the state of $Y$. Conversely, if $Y$ does not causally influence $X$, then the evolution of $X$ remains unaffected by the state of $Y$~\cite{ding_granger_2006}. Originally, Granger proposed methods to identify causality in linear systems using cross-spectral analysis. Since then, researchers have applied causal inference techniques in areas such as ecology and biology. More recently, there have been efforts to extend these methods to nonlinear systems, deal with noise in dynamics, and improve the efficiency of causal detection using fewer data points~\cite{hesse_use_2003,bressler_wienergranger_2011,bressler_wienergranger_2011,sugihara_detecting_2012,sinha_data-driven_2020,ancona_radial_2004,lungarella_methods_2007,ma_detecting_2014}.\vspace{-.2cm}

\section{Method}\vspace{-.2cm}
\subsection{Problem Statement}
This work addresses multi-vehicle interactions at roundabouts through a "learn-and-adapt" framework, where the learning and control phases are separated. The goal is to infer the safety specification of other vehicles through observation and use that knowledge to improve cooperative driving performance.
Given a set of demonstrations $\mathcal{D} = \{x_n^t, u_n^t\} \in \mathbb{R}^{N \times T}$, comprising state and control inputs from $N$ heterogeneous vehicles over $T$ time steps, our objectives are to:
\begin{enumerate}
    \item \textbf{Learn safety specifications:} Accurately infer the underlying safety specification $f(x, u, \alpha) \geq 0$ of the target vehicle from observational data, even in complex, multi-agent environments,  where it may exhibit different interactive behaviors toward different agents.
    \item \textbf{Design a safety-critical controller:} Based on the learned safety specification, develop a controller $u_i$ for the ego vehicle that enables safe and efficient cooperative driving.
\end{enumerate}

These objectives must address two core questions: \textit{(1) Which parts of the observational data should be used for learning?} and \textit{(2) How can the learned model be optimally leveraged to improve collective task performance?}

This paper relies on the following assumptions.
     First, {known safety radius ($R_{\text{safe}}$):} We assume that the safety radius for each vehicle is known. This is a reasonable assumption, as safety margins can often be estimated based on vehicle-specific features (e.g., size, braking capacity) and environmental conditions (e.g., road type, speed limits, human reaction time). For example, the National Safety Council’s ``3-Second Following Distance Rule'' provides a widely accepted guideline for safe driving~\cite{nsc}.   Second, {CBF-based controllers for other vehicles:} We assume that other vehicles are governed by CBF-controllers. While not all vehicles may implement exact CBF controllers, prior work~\cite{grover2022semantically} demonstrates that their behavior can often be approximated by CBF-based constraints. This assumption extends the applicability of our framework to a broader class of driving agents.  Third, {time-invariant behavior policies:} We assume each vehicle operates under a policy that remains invariant over time. This reflects human driving behavior, where underlying traits (e.g., aggressiveness or risk aversion) tend to remain stable, even if specific actions vary with context. By focusing on persistent behavioral characteristics, we simplify the learning and prediction tasks.

In the following sections, we introduce the Cross Map Smoothness-based causality reasoning algorithm in Sec. \ref{sec:cms-ci} and present a causality-embedded behavior prediction formulation in Sec. \ref{sec:behavior-prediction}. Finally, an impact-aware safety-critical control framework design is demonstrated in \ref{sec:controller-design} by leveraging the learned safety specification to achieve efficient collaborative driving. \vspace{-.3cm}

\subsection{CMS-based Causality Reasoning}\label{sec:cms-ci}\vspace{-.2cm}
The motivating questions for causality inference are: (1) "Did the target vehicle act that way because it wanted to (cost minimization) or because
it had to (safety constraint satisfaction)?” \cite{cosner2023learning}, and (2) "If safety constraints were the cause, which neighboring vehicle triggered the activation of those constraints?" 

To address these, we must identify when the target vehicle's safety constraint $f(x,u,\alpha)=\dot{h}(x,u) + \alpha H \geq 0$, particularly the Parametric CBF constraint \( \dot{h}(x,u) + \alpha H = 0 \), is activated, even without explicit knowledge of the safety behavior vector \( \alpha \). This involves pinpointing a subset of the demonstration $\mathcal{D}$ corresponding to times when the safety constraint is active. Recognizing that the activation of safety constraints indicates preventive actions by the target vehicle to maintain safety, we analyze the causal relationship between observed control signals of the target vehicle and its safety-related state evaluations during its interactions with surrounding vehicles.

We adopt the Cross Map Smoothness (CMS) method to assess the strength of causal relationships between two variables by comparing their time series. CMS is part of a wider family of algorithms known as Convergent Cross-Mapping, which is typically used in fields like genomics, biology, and ecology for detecting and quantifying causal interactions in complex, nonlinear systems~\cite{sugihara_detecting_2012, van_nes_causal_2015,deyle_predicting_2013, ma_detecting_2014, munch_nonlinear_2018}. The underlying theory behind it is that if variable ${X}$ causes $Y$, the states of $Y$ should be able to reconstruct $X$'s trajectory, demonstrating a smooth relationship between them over time. Our CMS assessment incorporates three key components: ``time-delayed observations," ``delay embedded vectors," and "reconstructed attractors"~\cite{rand_detecting_1981}. We retain standard terminology from causality inference literature and provide definitions in our study's context.

For each pairwise relationship between the target vehicle $j$ and other surrounding vehicles $k\in N \setminus j$, we denote "time-delayed observations" \cite{ma_detecting_2014} of the control signal and states at time $t$ as $u_j(t)$ and $h_{jk}(t)$. $u_j(t)$ is the \( \ell^2 \)-norm of the observed control signal $u$, i.e., observed velocity, of the target vehicle at time $t$. $h_{jk}(t) = \lVert x_j^t-x_k^t \rVert^2-R_{safe}^2$ is the safety function characterizing the desired safe distance to maintain the pairwise inter-vehicle safety, calculated with the states of the target vehicle $j$ and its pairwise companion $k$.

For a given receding horizon $L$, delayed-embedding vectors are therefore defined as
\begin{equation}\footnotesize
    \bold{u}_j(t)=\begin{bmatrix}
        u_j(t) \\
        u_j(t-1) \\
        \vdots \\
        u_j(t-L-1)
    \end{bmatrix},  \bold{h}_{jk}(t)=\begin{bmatrix}
        h_{jk}(t) \\
        h_{jk}(t-1) \\
        \vdots \\
        h_{jk}(t-L-1)
    \end{bmatrix}
\end{equation}
which is a collection of past observations in reverse chronological order.
For the entire observation horizon $T$, the reconstructed attractors aggregate the delayed-embedded vectors together, denoted as $M_u=\{\bold{u}_j(t), \bold{u}_j(t-1), ... \bold{u}_j(t-T)\}$ and $M_{h}=\{\bold{h}_{jk}(t), \bold{h}_{jk}(t-1), ... \bold{h}_{jk}(t-T)\}$ respectively. 

According to the definition of causality, if $u_j(t)$ causally influences $h_{ij}(t)$, then the state of $u_j(t)$ is a driving factor in the dynamics of $h_{jk}(t)$, meaning that information about $u_j(t)$ is reflected in $h_{jk}(t)$'s dynamics. Conversely, if $u_j(t)$ does not causally influence $h_{jk}(t)$, then the dynamics of $h_{jk}(t)$ remain insensitive to changes in $u_j(t)$. The construction of reconstructed attractors $M_u$ and $M_h$ from time series $u_j(t)$ and $h_{jk}(t)$, respectively, is crucial to evaluate the extent to which the state information of $u_j(t)$ is embedded within the dynamics of $h_{jk}(t)$. Specifically, we assess the smoothness of the mapping between $M_u$ and $M_h$. If there exists a smooth cross map from $M_h$ to $M_u$, it suggests that $u_j(t)$ causally influences $h_{jk}(t)$.

We quantify the smoothness of this cross map using the error of a Radial Basis Function Neural Network (RBFNN) \cite{ma_detecting_2014}, which is tasked with learning the mapping between attractors. Since RBFNNs are capable of approximating any smooth function, a low error implies a smooth mapping and thus strong causality between $u_j(t)$ and $h_{jk}(t)$. Conversely, a high error suggests that the mapping is not smooth, indicating little to no causality between the two variables. Readers are referred to~\cite{rand_detecting_1981, sugihara_detecting_2012, ma_detecting_2014} for more details.

Here we present our CMS-based Causality Reasoning Algorithm described in Algorithm \ref{alg:CMS}. For brevity, we refer to $\bold{u}(t-s)$ as $\bold{u}_s$ and $\bold{h}(t-s)$ as $\bold{h}_s$. 
\vspace{-0.5cm}
\begin{algorithm}
\caption{\footnotesize Cross Map Smoothness-based Causal Reasoning } \footnotesize\label{alg:CMS}
\KwData{$\bold{h_1, h_2, ..., h_n} \in \mathbb{R}^L$ and $\bold{u_1, u_2, ..., u_n} \in \mathbb{R}^L$, let $S_m=\{1,2,...,n\}\backslash m$ be the leave-one-out index set.}
\KwResult{Causality Index: $R_{hu} \in [0,1]$}
\For{$m=1,2,...,n$}{
    train radial basis function neural network $\mathcal{N}_m$ based on $\mathcal{N}_m(\bold{u}_s)=\bold{h}_s, s\in S_m$ \\
    $\hat{\bold{h}}_m\leftarrow \mathcal{N}_m(\bold{u}_m)$ \\
    $\epsilon_m\leftarrow \|\bold{h}_m-\hat{\bold{h}}_m\|$
}
Normalize the error: $\Delta \leftarrow \frac{\texttt{rms}(\bold{\epsilon})}{\texttt{rms}(\|\bold{h}-\Bar{\bold{h}}\|)}$ \\
\tcc{$\sigma$ is a positive constant used to normalize $R_{hu}$.} 
Causality Index:  $R_{hu}\leftarrow \frac{1}{\texttt{exp}(\Delta/\sigma)}$ 
\end{algorithm}
 The input to the algorithm is then the corresponding sets of time-delayed coordinate vectors of size $n$: $\{\bold{h}_1, \dots, \bold{h}_n\} $ and $ \{\bold{u}_1, \dots, \bold{u}_n\}$. These sets are the reconstructed attractors described above. Leave-one-out-sets $S_m=\{1,2,...,n\}\backslash m$ are then used to train an RBFNN $n$ times to determine the network error, which is inversely proportional to the causality index. The output of the CMS algorithm is a continuous scale from 0 to 1 quantifying various levels of causality, where 1 indicates strong causal influence and 0 indicates none.

\subsection{Behavior Modeling with CMS-enabled Parametric CBF}\label{sec:behavior-prediction}
Upon establishing paired dynamics, we employ linear ridge regression to predict safety behaviors. The CMS-based causality reasoning offers a key advantage over existing approaches like \cite{lyu_adaptive_2022} by relaxing the assumption that the exact activation instances of the CBF constraint must be known. With CMS, we can accurately detect when the vehicle's constraint is inactive, that is, when the CBF constraint 
$f(x,u,\alpha)\geq 0$ is not enforced. This flexibility allows us to transform the original constrained optimization problem, as shown in Eq. \ref{eq:constrained-optimization}, into an unconstrained optimization problem. In this new formulation, constraint satisfaction is managed by CMS-based causality inference, which selectively uses observations of $h_{jk}^t$ corresponding to time instances when the constraint is identified as active, facilitating accurate parameter estimation: \vspace{-.5cm}
\begin{equation}\footnotesize
\begin{aligned}
\Bar{\alpha}_j&=\underset{\alpha_j}{\arg\min}\sum^m_{t=1}\|\Dot{h}^t_{jk}-\alpha_j H^t_{jk}\|^2_2+r\|\alpha_j\|^2_F \\
        &\text{s.t.} \quad \Dot{h}_{jk}^t(x,u)+\alpha_j H^t_{jk}(x) \geq0
        \label{eq:constrained-optimization}
\end{aligned}
\end{equation}
    where $r$ is the regularizer parameter, $\|\alpha_j\|_F$ is the Frobenius norm of $\alpha_j$, and $\Bar{\alpha}_j$ is the estimated solution. 
    
To address the noise in observations, we leverage the theoretical properties of the class 
$\kappa(\cdot)$ functions and the forward invariance of CBFs to impose additional bounds on the parameter estimation to characterize those likely invalid inference results, tightening the estimation for better accuracy. Specifically, we introduce:
1) Monotonicity Condition: By definition, the function $\kappa(h(x))=\alpha H(x)$ must be strictly increasing, which requires all elements of 
${\alpha}$
to be non-negative. Thus, if
$\min\Bar{\alpha}<0$,  this estimate of $\alpha$ is invalid.
2) Constraint Satisfaction: The inequality $\Dot{h}(x,u)+\alpha H(x)  + \delta_c\geq0$ must hold for the true parameter, where $\delta_c$ is a small value very close to 0 to allow for slackness under noisy motion. Therefore, if the left hand side is less than 0,  this estimate of $\alpha$ is also invalid. 
3) Interaction Condition: For two vehicles to be considered interacting, the constraint should remain close to zero, indicating a near-boundary condition of the safety set. 4) Temporal Consistency: we impose a fourth condition that the current estimate aligns closely with previous estimates, under the assumption that the safety behavior parameter of the vehicle remains time-invariant.  These conditions help refine parameter estimates and maintain robustness against noise, ensuring that only feasible solutions are retained. After identifying and removing those invalid estimates, the rest of the estimation of $\alpha$ is stored in a dataset $\mathcal{D}_{jk}$ to be used to obtain the final estimate value by averaging across this dataset. The pseudocode of the proposed CI-enabled Parametric CBF estimation is described in Algorithm \ref{rpe}. In lines 1-6, we use CMS to test if there is an active constraint between all pairs of vehicles using Algorithm \ref{alg:CMS}. If the causality index is detected to be above a certain threshold, we begin the safety behavior estimation algorithm in line 7. A typical mean squared error minimization problem (Eq. \ref{eq:constrained-optimization} without constraints) can be solved analytically.
\begin{equation} \footnotesize
    -\begin{bmatrix} H^T(t_1) \\ \vdots \\ H^T(t_m) \end{bmatrix}^T \begin{bmatrix} H^T(t_1) \\ \vdots \\ H^T(t_m) \end{bmatrix} \alpha = \begin{bmatrix} H^T(t_1) \\ \vdots \\ H^T(t_m) \end{bmatrix}^T \begin{bmatrix} \Dot{h}(t_1) \\ \vdots \\ \Dot{{h}}(t_m) \end{bmatrix}
    \label{eq:inv}
\end{equation}

Since we are unsure if the surrounding vehicles actually take actions that maintain their safety behavior constraints, we opt for an online sequential solver that satisfies the 4 conditions mentioned above instead of performing the large matrix inverse operation required to solve for $\alpha_j$ in Eq. \ref{eq:inv}. Lines 9-11 estimate a new $\alpha_j$ based on the current $H_{jk}$ and $\Dot{h}_{jk}$ values. In line 12, estimation validity is verified through the additional bounds we introduced based on the monotonicity condition and constraint satisfaction.
Line 16 considers valid estimates of $\alpha_j$ and checks for interaction activation ($\hat{f}(x,u,\alpha_{new})\approx 0$) and temporal consistency ($\epsilon < \delta_{rmse}$). If the valid estimate passes these conditions as well, it is added into the dataset. Averaging the values in the dataset returns the final estimated value of $\alpha_j$, concluding the behavior modeling of the target vehicle.

\vspace{-.4cm}
\begin{algorithm} \footnotesize
\caption{\footnotesize Causality-embedded CBF Parameter Estimation} \label{rpe}
\KwData{$\Delta x_{jk}, \Delta v_{jk}, u_j, u_k, \Delta t, R_{safe}$}
\KwResult{$\Bar{\alpha}_j$}

\While{True}{
    Add $\{x^t_n, u^t_n\}$ to $\mathcal{D} \forall n \in N$ 
    \uIf{$t>L$}{
        For each vehicle pair (j, k): $\mathcal{R}_{hu}=\text{CMS}(\bold{h}_{t-L-1} ..., \bold{h}^t, \bold{u}_{t-L-1}, ..., \bold{u}^t)$ 
        \uIf{$\mathcal{R}_{hu}>\epsilon_{R}$}{
            Jump to line 7 
        }
    }
}
Initialize $A_{sum}$ and $B_{sum}$ to $0$ 
\For{$t=T_1:T_2$}{
    Calculate $A^t \leftarrow H_{jk}^t{H_{jk}^t}^T$ and $B^t\leftarrow H_{jk}^t\Dot{h}_{jk}^t$ 
    Update $A_{sum} \leftarrow A_{sum} + A^t$ and $B_{sum} \leftarrow B_{sum} + B^t$ 
    Estimate $\alpha_{new} \leftarrow [-A_{sum}^{-1}B_{sum}]^T$ 

    Calculate estimated constraint: $\hat{f}(x,u,\alpha_{new})\leftarrow\Dot{h}_{jk}^t(x,u_j)+\alpha_{new} H^t_{jk}(x)$ 
    
    $\epsilon_{rmse} \leftarrow$ \texttt{rmse}$(\alpha_{new}, \alpha_{old})$ 

  \uIf{$|\hat{f}(x,u,\alpha_{new})|>\delta_c$ or $\min\alpha_{new}<0$}{
    Reinitialize $A_{sum}$ and $B_{sum}$ to $A^t$ and $B^t$
  }
  \ElseIf{$\hat{f}(x,u,\alpha_{new})\approx 0$ and $\epsilon<\delta_{rmse}$}{
    Add $\alpha_{new}$ to dataset $\mathcal{D}_{jk}$ 
    }{\texttt{$\alpha_{old}\leftarrow\alpha_{new}$} 
  }
} 
$\hat{\alpha}_j \leftarrow$ average($\mathcal{D}_{jk}$) %
\end{algorithm}
\vspace{-.7cm}

\subsection{Impact-Aware Control Strategy}\label{sec:controller-design}
In this section, we demonstrate how to leverage the learned behavior of the target vehicle to enhance collaborative driving. Specifically, our objective is to improve collective efficiency with smooth driving interactions by maximizing the comfort of the target vehicle’s motion and minimizing any jerkiness induced by the presence of the ego vehicle. We present our approach as the following optimization problem, and we refer to the ego vehicle with index $i$ and the target vehicle with index $j$: 
\begin{equation}\footnotesize
    \begin{aligned}
        u_i^t = & \underset{u_i}{\arg\min} \quad c_1||u_i^t-\Bar{u}_i^t||^2  \\
        & + c_2 \min\{\Dot{h}_{ij}^{t_f}(x,u)+\bar{\alpha}_j H_{ij}^{t_f}(x), 0\}\Dot{h}_{ij}^t(x,u) \\
        \text{s.t.} \quad & \Dot{h}_{ij}^t(x,u)+\alpha_i H_{ij}^t(x)\geq0 \\
    \end{aligned}
\end{equation}

where the output $ u_i^t$ represents the actual control signal of the ego vehicle at the current timestep $t$ and $\bar{\alpha}_j$ represents the learned safety behavior. The time derivative  $\Dot{h}_{ij}^{t_f}(x,u)$ and parametric CBF $H_{ij}^{t_f}(x)$ are calculated at the future timestep $t_f$. We argue that these values can be reasonably calculated due to the advances in learning models for vehicle behavior prediction \cite{mozaffari_deep_2022}. In addition, for our experiments, we demonstrate that even a simple piece-wise constant velocity model, $x^{t_f}=x^t + (t_f-t)\Dot{x}^t$, can suffice.

The objective function consists of two positive terms with $c_1$ and $c_2$ as corresponding coefficients that represent relative priority. The first term instructs the system to follow the nominated control input $\bar{u}_i^t$. The second term maximizes the current $\Dot{h}^t_{ij}(x,u)$ according to the severity of the learned future safety violation $f(x^{t_f}, u^{t_f}, \bar{\alpha})=\Dot{h}_{ij}^{t_f}(x,u)+\bar{\alpha}_j H_{ij}^{t_f}(x)$. If the constraint is not violated, i.e., $f(x^{t_f}, u^{t_f}, \bar{\alpha}_j)\geq0$, then the second term becomes 0 due to the $\min\{ \cdot, 0\}$ operator. Conversely, if the constraint is violated, then $c_2 \min\{\Dot{h}_{ij}^{t_f}(x,u)+\bar{\alpha}_j H_{ij}^{t_f}(x), 0\}$ is a negative multiplier, incentivizing $\Dot{h}_{ij}^t(x,u)$ to be maximized in the optimization problem. Greater violations are reflected in greater negative multipliers, resulting in higher relative priorities to maximize $\Dot{h}_{ij}^t(x,u)$, thus minimizing the impact the ego vehicle has on the target.

\vspace{-.5cm}
\section{Experiments and Discussion}

In this section, we first present the performance of the CMS-enabled Parametric CBF for behavior modeling in multi-vehicle interaction scenarios. Then we demonstrate how collaborative efficiency could be improved via our impact-aware control strategy design.

\subsection{CMS-enabled Parametric CBF for Behavior Modeling}
\subsubsection{Causality Detection in Multi-Vehicle Interactions}

The scenario we selected is a 7-vehicle roundabout, as shown in Fig. \ref{fig:CMS} (left), which represents a typical challenging urban driving scenario due to the need for intensive inter-vehicle interactions. The ego vehicle’s objective is to navigate the roundabout safely while optimizing for collective efficiency. In this single-lane setting, the ego vehicle must focus on safely merging by interacting with a single target vehicle. However, no prior information is available to predict the target vehicle’s movements. Therefore, the ego vehicle must model the target’s behavior by observing its interactions with neighboring vehicles and inferring its safety specifications. We denote the seven vehicles, excluding the ego vehicle, as $s1, \dots, s7$.

We apply our proposed CMS-based causality reasoning algorithm to this 7-vehicle interaction scenario, with results shown in Fig. \ref{fig:CMS}
 (right). Using noisy observations of all vehicles' past trajectories as inputs, the algorithm performs causality detection for each vehicle pair involving the target vehicle $s_5$ and all other vehicles, illustrating the strength of cause-and-effect relationships that evolve over time. As expected, not all surrounding vehicles influence the movement of the target vehicle. In fact, as indicated by the CMS value plots, only $s_1$ and $s_7$ show a strong causal relationship with 
$s_5$’s behavior during specific time periods, indicating that $s_5$'s safety constraint is activated only between these vehicle pairs during these particular periods.

\subsubsection{Causality-Embedded Behavior Modeling}

\begin{table*}[h!]
    \centering
    \begin{tabular}{ |p{3cm}||p{3cm}|p{3cm}|p{3cm}|p{3cm}|  }
        \hline
        \multicolumn{5}{|c|}{Trial Results} \\
        \hline
        Method&  Average MSE (all) &Std Dev. MSE (all)&Average MSE ($95\%$) &Std Dev. MSE ($95\%$)\\
        \hline
        Ours w/CMS  & 0.3267 & 3.1323  & 0.0023   & 0.0041 \\
        NBF w/CMS & 0.6905 & 3.7423 & 0.1678 & 0.1477 \\
        Parametric-CBF \cite{lyu_adaptive_2022} & 0.3692 & 0.2121 & 0.3454 & 0.1893 \\
        NBF w/o CMS & 0.5832 & 0.2754 & 0.5531 & 0.2481\\
        \hline
    \end{tabular}
    \caption{\footnotesize Accuracy comparison with existing approaches}
    \label{compare}
\end{table*}

\begin{figure*}[ht]
    \centering
    \begin{subfigure}[t]{0.5\textwidth}
        \centering
\includegraphics[width=\textwidth]{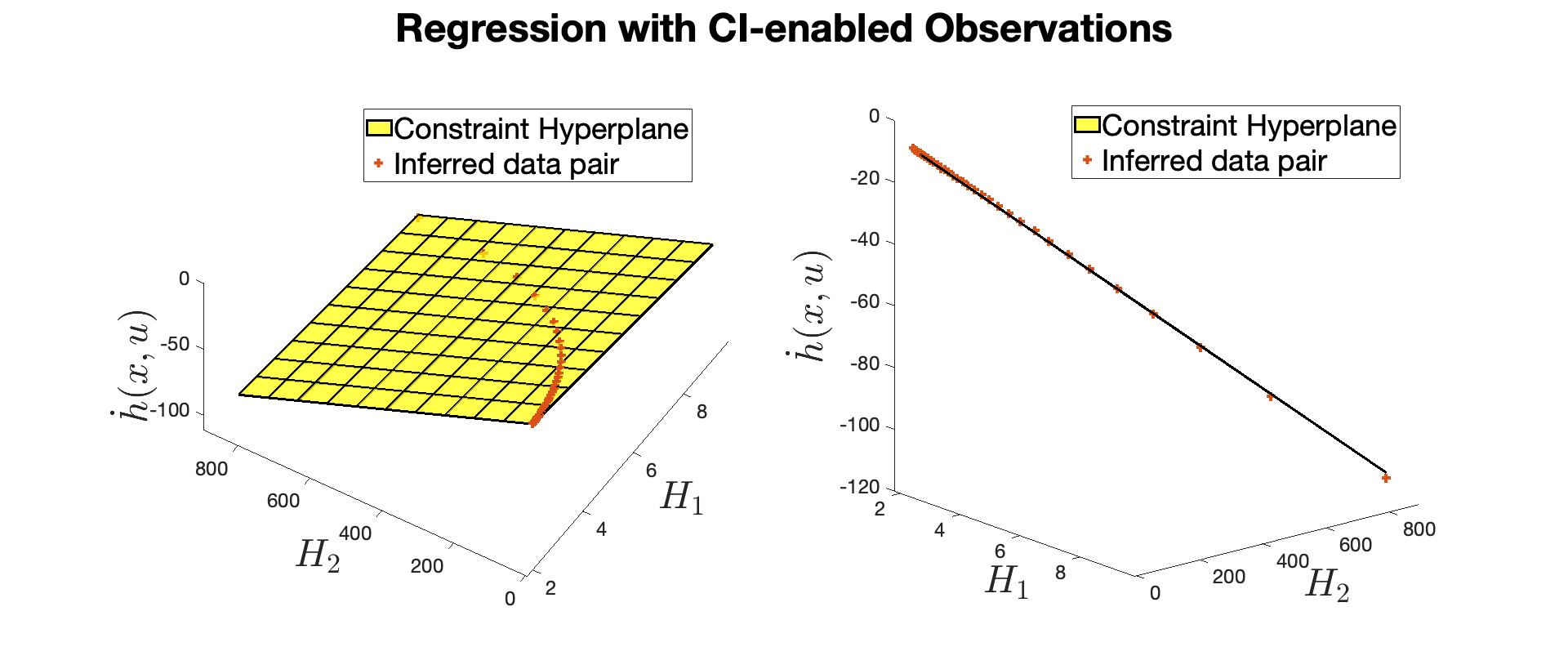}
        \caption{\footnotesize Learned constraint hyperplane with inferred causal data pair}
        \label{fig:constraint-plane}
    \end{subfigure}
    \hfill
    \begin{subfigure}[t]{0.48\textwidth}
        \centering     \includegraphics[width=\textwidth]{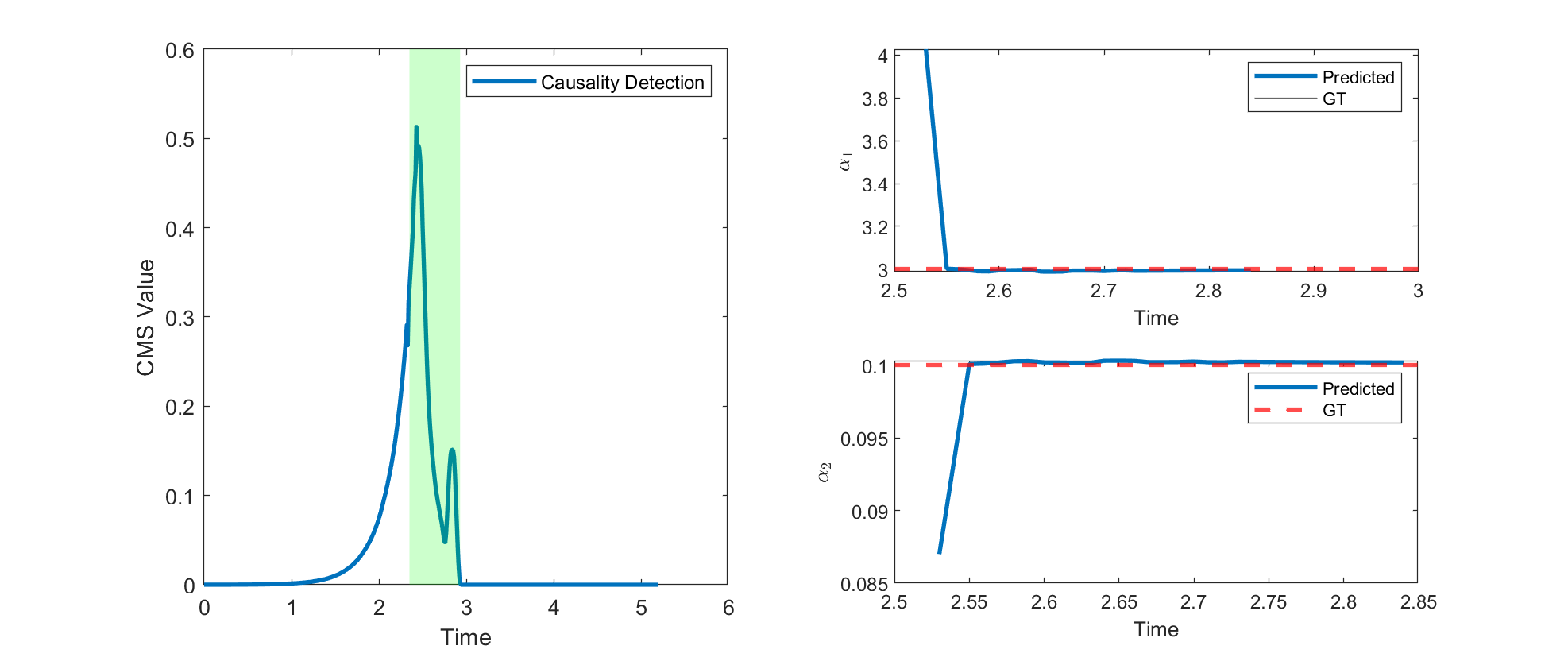}
        \caption{\footnotesize Learned Causality-based Parametric-CBF}
        \label{fig:enter-label}
    \end{subfigure}
    \caption{\footnotesize CI-enabled Constraint Learning.}
    \label{fig:side-by-side}
\end{figure*}

Using the CMS-based causality inference algorithm, we identify the most relevant portion of data from the full set of observations and supply this selective subset to our parametric CBF estimation procedure. Figure~\ref{fig:constraint-plane} illustrates this process: in the left panel, the red markers represent the inferred data pairs from an example trial, while the yellow grid shows the learned constraint hyperplane. The close alignment of the markers with the hyperplane indicates that the CMS-selected samples are consistent with the estimated constraint and thus faithfully capture the causal structure of the interaction. In the right panel, the same data and hyperplane are shown edge-on, further validating this consistency by demonstrating that all inferred data pairs lie directly on the hyperplane. The learned hyperplane encodes the target vehicle’s underlying safety specification, effectively characterizing the boundary conditions that govern its behavior. This representation offers an interpretable and compact model of how the vehicle enforces safety in the observed scenario. 
Then we demonstrate the safety specification learning for behavior modeling in Fig. \ref{fig:enter-label}. By identifying the portion of data with causality marked in a green background, we show on the right that our CI-enabled Parametric CBF is able to quickly converge to the ground truth values marked as red dashed lines.

To further emphasize the effectiveness of our approach, we compare it to a popular class of approaches for learning control barrier function parameters \cite{cosner2023learning}, \cite{qin_learning_2021}, \cite{yang_enhancing_2024}, \cite{robey_learning_2020}, which turns CBF hard constraints into soft constraints by reformulating them into the objective function. The original parameter learning problem is as follows, where functions $h(x)$ and $\kappa(h(x))$ are parameterized by $\theta$ and $\alpha$. 
\begin{equation} \footnotesize
    \label{eq: og equation}
    \begin{aligned}
        \alpha^*, \theta^* = & \underset{\alpha, \theta}{\min} \quad |\theta| + |\alpha| \\
        \text{s.t.} \quad & \Dot{h}_{\theta}(x,u)+\kappa_\alpha (h_\theta(x))\geq0 \quad \forall \; (x,u) \in \mathcal{D}_\text{safe} \\
        & h_\theta(x)\geq0 \quad \forall \; (x,u) \in \mathcal{D}_\text{safe} \\
        & \Dot{h}_{\theta}(x,u)+\kappa_\alpha (h_\theta(x)) <0 \quad \forall \; (x,u) \in \mathcal{D}_\text{unsafe} \\
        & h_\theta(x) <0 \quad \forall \; (x,u) \in \mathcal{D}_\text{unsafe}
    \end{aligned}
\end{equation}
Here, $\mathcal{D}_{safe}$ and $\mathcal{D}_{unsafe}$ represent $(x,u)$ pairs that satisfy $h(x)>0$ for the safe set and $h(x)<0$ for the unsafe set. Conditioning on $h_\theta (x)$ being known and $\kappa_\alpha(h(x))=\alpha H(x)$, \cite{cosner2023learning}, \cite{qin_learning_2021}, \cite{yang_enhancing_2024}, \cite{robey_learning_2020} reformulated the optimization problem as: $\alpha^* =  \underset{\alpha}{\min} \quad |\alpha| + \sum_{(x,u) \in \mathcal{D}_\text{safe}} \max (0, -(\Dot{h}_{\theta}(x,u)+ \alpha H(x))$. {Since this optimization loss is typically used to train Neural Barrier Functions, we address this method as NBF in Table \ref{compare}.}

\begin{figure*}
    \centering
    \begin{subfigure}{0.28\textwidth}
        \includegraphics[width=\linewidth]{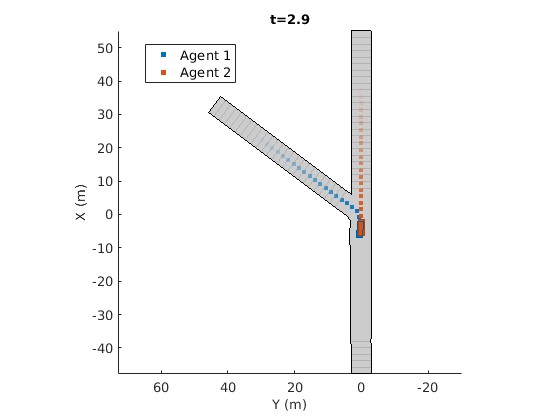}
        \caption{\footnotesize Collision due to infeasibility of conventional CBF-QP}
    \end{subfigure}
    \begin{subfigure}{0.28\textwidth}
\includegraphics[width=\linewidth]{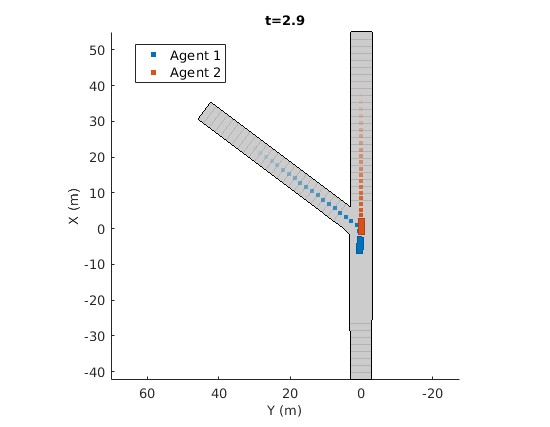}
\caption{\footnotesize Safe merging with our impact-aware control}
    \end{subfigure}
        \begin{subfigure}{0.4\textwidth}
\includegraphics[width=.8\linewidth]{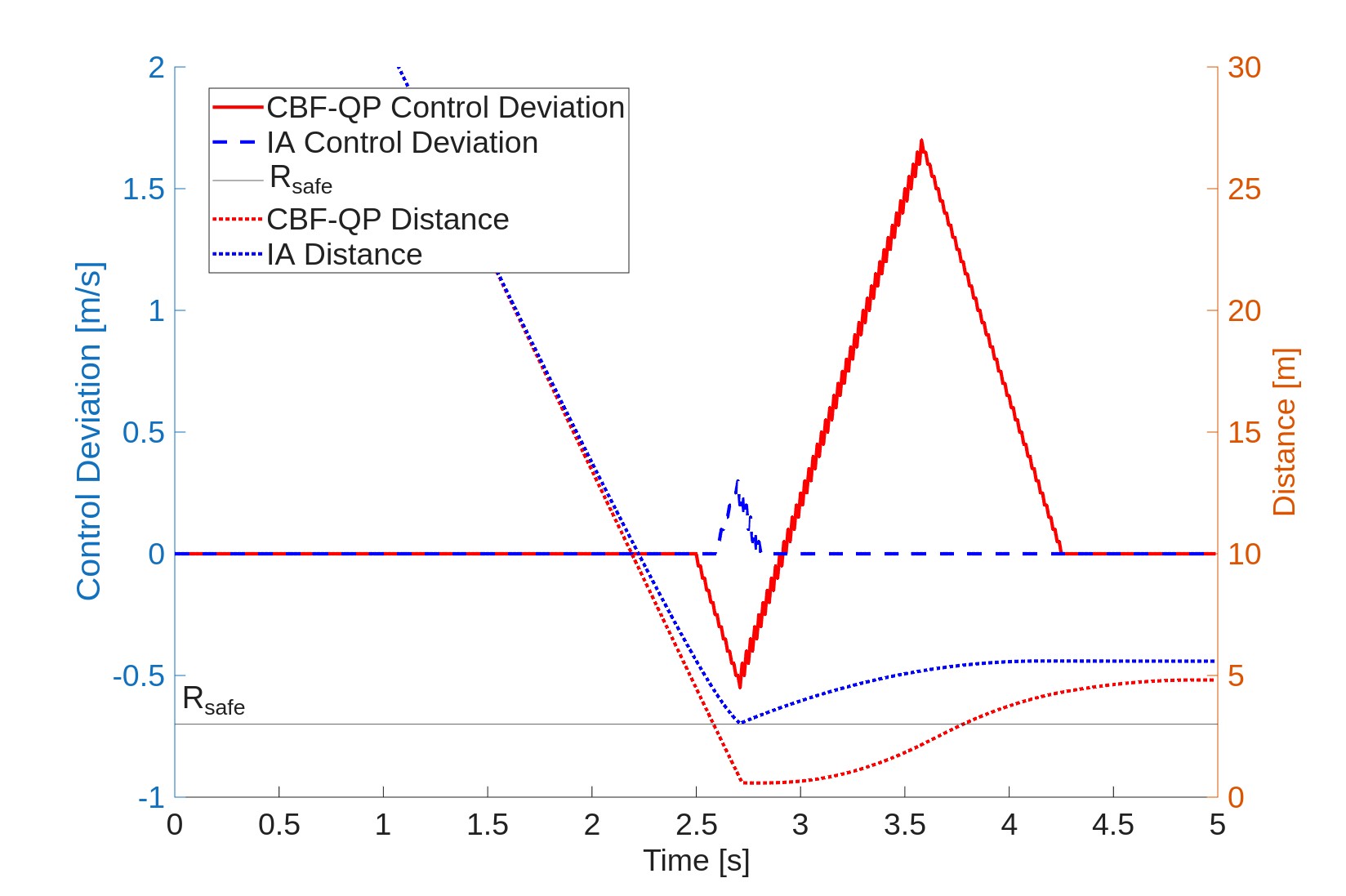}
\caption{\footnotesize Control deviation (left) and inter-vehicle distance (right) over time}
    \end{subfigure}
    \label{fig:IA}
    \caption{\footnotesize Performance comparison in ramp merging}
\end{figure*}
\begin{figure*}
    \centering
    \begin{subfigure}{0.26\textwidth}
        \includegraphics[width=\linewidth]{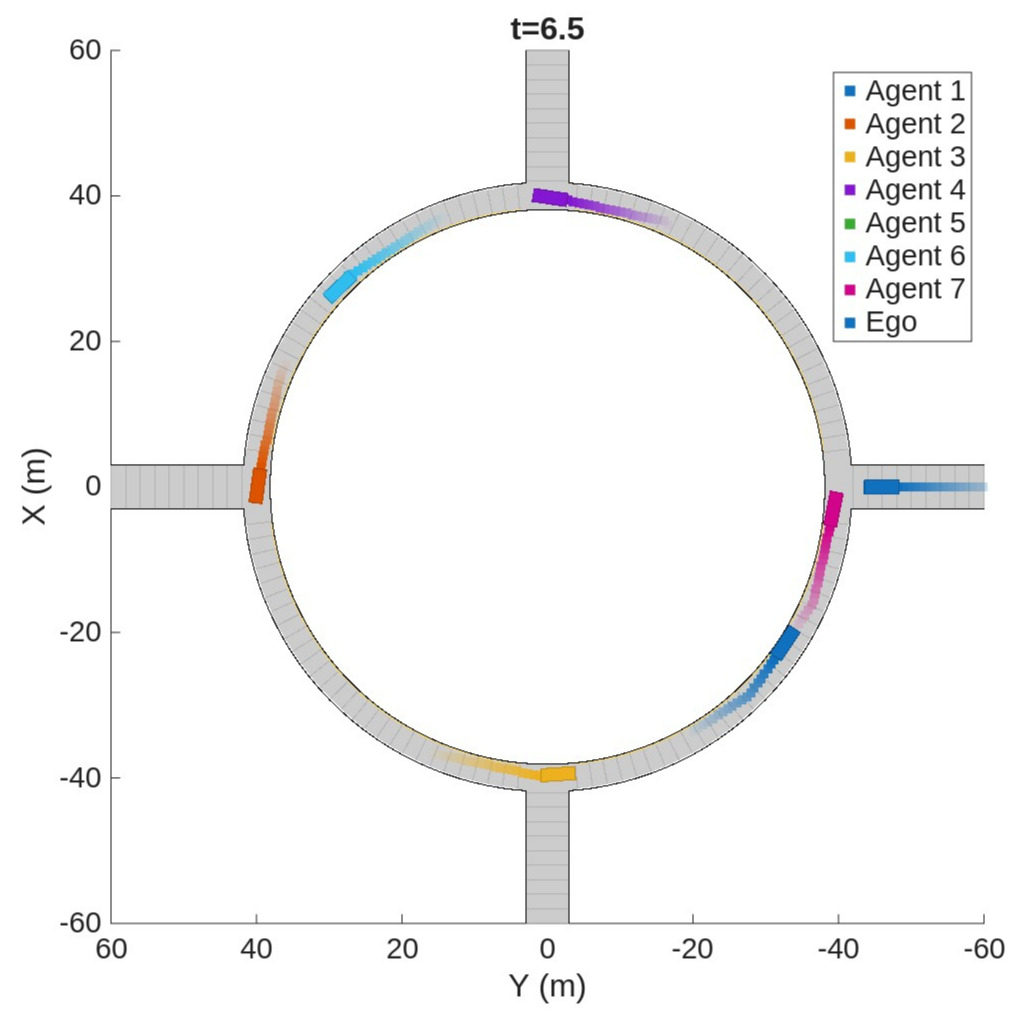}
        \caption{\footnotesize Deadlock due to over conservativeness of conventional CBF-QP}
    \end{subfigure}
    \begin{subfigure}{0.26\textwidth}
\includegraphics[width=\linewidth]{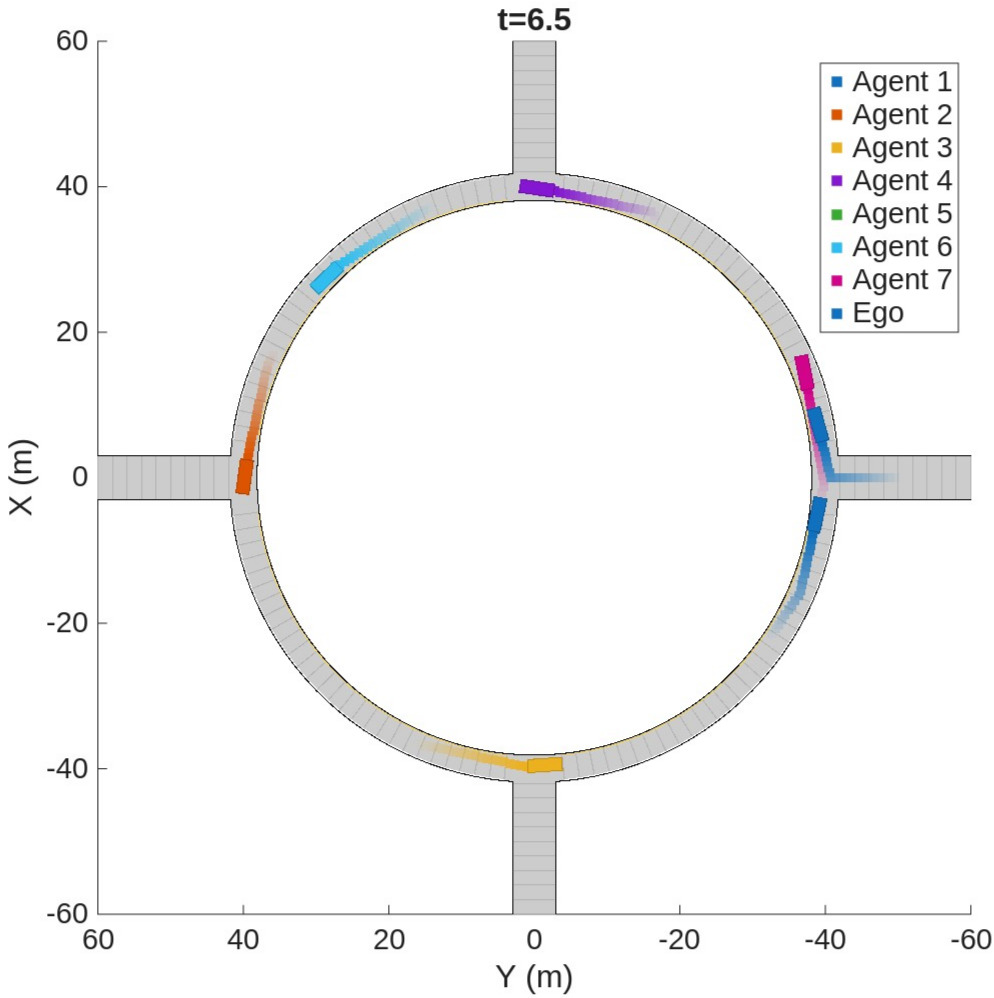}
\caption{\footnotesize Safe merging between Agents 1 and 7 with our impact-aware control}
    \end{subfigure}
        \begin{subfigure}{0.46\textwidth}
\includegraphics[width=\linewidth]{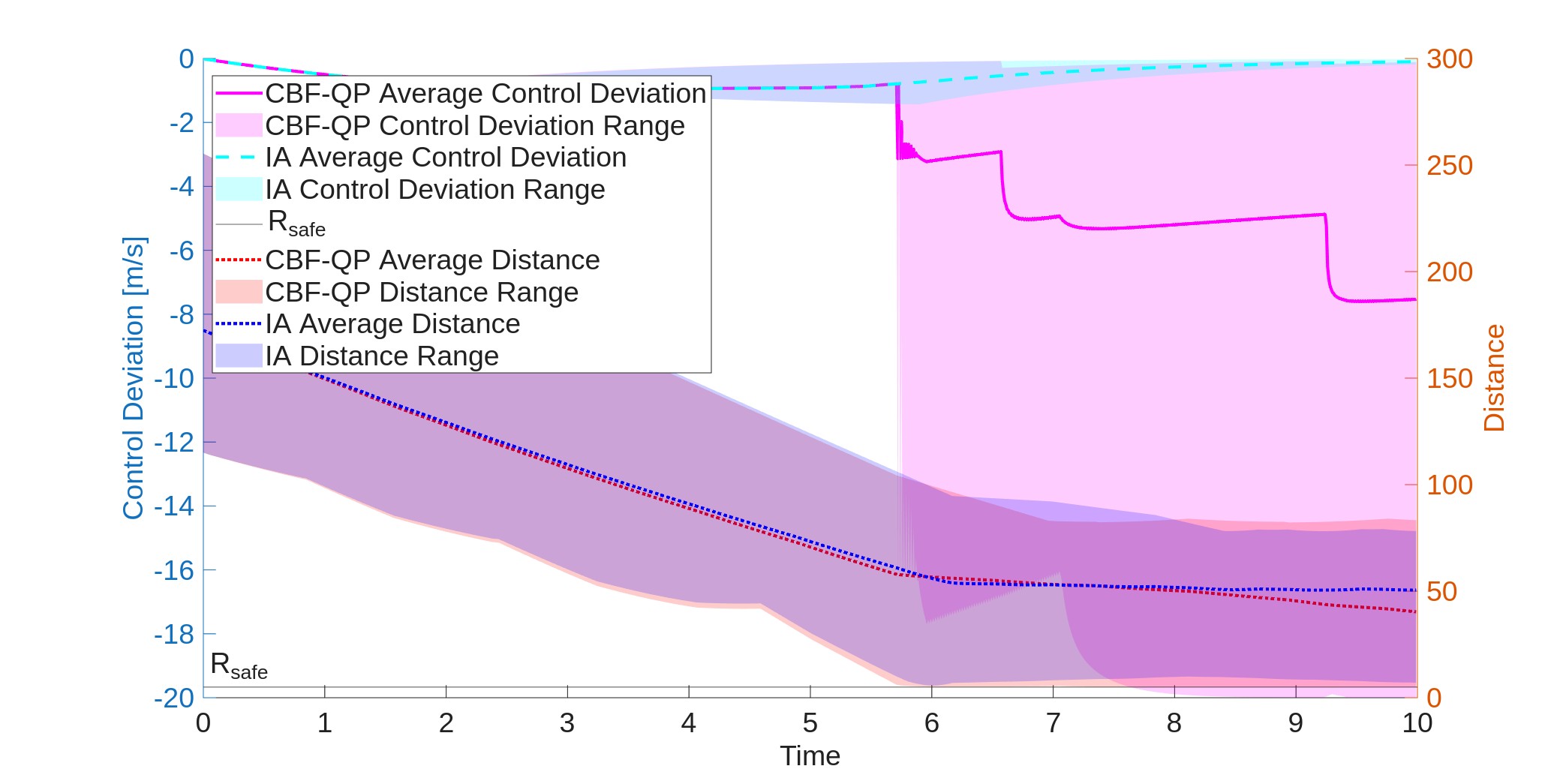}
\caption{\footnotesize Control deviation (left) and inter-vehicle distance of all vehicles (right) over time}
    \end{subfigure}
    \label{fig:round-IA}
    \caption{\footnotesize Performance comparison in roundabout merging}
\end{figure*}\vspace{-.4cm}
We compare the results of their solution with our proposed method as shown in Algorithm \ref{rpe} in the roundabout scenario. In this scenario, the target vehicle interacts with other vehicles, and the ego vehicle perceives these interactions through noisy observations. However, no information is provided regarding when and which vehicles are actively interacting. Over 200 trials, with motion noise sampled from $\mathcal{N}(0,1)$, the target vehicle uses different values of $\alpha$ to induce varying safety behaviors.

Table~\ref{compare} presents the results of the two approaches. The left two columns show the average and standard deviation of the mean squared error (MSE) between the true $\alpha$ value and the predicted value across all 200 trials for both methods. The right two columns report the same statistics computed over the $95\%$ of trials with the lowest MSE. This additional metric highlights that Algo.~\ref{rpe} performs exceptionally well in the majority of cases. However, there are rare instances where the algorithm's performance degrades, due to fixed values of $\delta_c$ and $\delta_{rmse}$ that are not tuned for individual trials. In comparison, the original ridge regression method \cite{lyu_adaptive_2022} demonstrates consistent performance across all trials, including the $95\%$ subset, but does not consistently achieve lower error due to the presence of observation noise. From the comparison results, we conclude that our approach outperforms existing methods, with the effectiveness of CMS demonstrated by a significantly higher average accuracy across the majority of test cases.

Our approach achieves superior performance primarily due to two key factors: (1) strong robustness to localization uncertainty in the observed data, and (2) the ability to identify data segments with a strong causal relationship to the safety constraint being learned. While all agents aim to act safely, motion uncertainty can lead to interactions that violate safety constraints, as verified by the Control Barrier Function conditions. Existing methods typically rely on clear labels to distinguish safe from unsafe data. However, in uncertain settings, such labels are often unavailable or unreliable—leading these methods to treat all observations as safe, which compromises robustness in parameter learning. Even when labels are available, these methods fail to isolate the truly informative segments of data, leading to overestimation compared to ground truth values. In contrast, our approach leverages causal inference to pinpoint portions of data where safety constraints are actively engaged. This targeted learning results in higher overall accuracy.

\subsection{Impact-Aware Control}
We argue that our proposed impact-aware control has two advantages: 1) improving the collaborative driving efficiency, and 2) enhancing the collective safety in these interactions. To demonstrate the effectiveness of our approach, we validated it on two scenarios: ramp-merging on highway driving (one-on-one interaction) and a roundabout in urban driving (multi-agent interactions, 7 agents in total excluding ego). {In both scenarios, 100 trials were conducted with ego vehicles starting from uniformly sampled positions away from the merging point (between 38 and 46m). In the ramp scenario, the other vehicle is also uniformly sampled from the same range, while the roundabout scenario has surrounding vehicles starting in preset positions. These surrounding vehicles are simulated with conventional CBF-QP safe controllers with a nominal speed of {20m/s}. For evaluation, we analyze the following metrics: the total time to complete the collaborative driving task, the speed deviation from the nominal controllers, and the inter-vehicle distance over time.}

{In the ramp merging scenario, the mean and standard deviation of the total time to complete the driving task over 100 trials are {3.42s} and {1.34s} with our proposed approach, while the mean and standard deviation using the conventional CBF-QP method are {4.24s} and {1.54s}. 
In the roundabout scenario, the mean and standard deviation completion times are {9.77s} and {3.60s} with our approach, while the CBF-QP yielded {10.26s} and {3.83s}.}
The {19.34\% and 4.87\%} improvement in task efficiency is primarily due to the impact-aware design of our method. By promoting more courteous behavior, the ego vehicle not only optimizes for its own objectives but also considers the potential disruption it may cause to other agents, thereby minimizing overall control deviation. This design becomes particularly advantageous in more complex scenarios, where mutual consideration greatly improves coordination and system performance. As shown in Fig. \ref{fig:IA}(c), during one-on-one interactions, our approach causes significantly less disruption to the other vehicle’s trajectory compared to the conventional CBF-QP, avoiding abrupt reactions and enabling smoother interaction.

In some trials, the conventional CBF-QP executed by other vehicles may yield infeasible solutions due to noisy observations. The standard fallback strategy in such cases is to apply an emergency brake and come to a full stop, as illustrated in Fig. \ref{fig:IA}(a). In contrast, our impact-aware approach enables the ego vehicle to anticipate and mitigate these situations. By explicitly quantifying the feasibility space in the near future of the other vehicle using our learned Parametric CBF, the ego vehicle can proactively plan preventative actions with lookahead capabilities when a potential infeasibility is foreseen, as shown in Fig. \ref{fig:IA}(b). This advantage is further supported by Fig. \ref{fig:IA}(c), where the inter-vehicle distance under the conventional CBF-QP dips below the safety threshold (black dashed line), indicating a safety violation. Similar results are observed in the roundabout scenario, as shown in Fig.~\ref{fig:round-IA}. In Fig.~\ref{fig:round-IA}(a), using the conventional CBF-QP, a deadlock occurs because both the ego vehicle and Agent~7 plan their actions independently, without accounting for each other. In contrast, Fig.~\ref{fig:round-IA}(b) demonstrates our impact-aware design, where the ego vehicle proactively considers the admissible action space of both parties. By accelerating to reduce ambiguity, it nudges Agent 7 forward, creating sufficient space for itself while avoiding danger and preventing deadlock. This results in a smoother collaborative outcome, where all vehicles continue progressing rather than being forced to stop. Fig.~\ref{fig:round-IA}(c) shows that, in comparison,  our approach consistently maintains a safe distance above the margin, improving the collective safety, while causing much less control deviation for others, reflecting a more courteous and efficient behavior paradigm.

\vspace{-.3cm}
\section{Conclusion}
\vspace{-.2cm}
In this work, we propose a novel Causality Inference-enabled Parametric Control Barrier Function framework for safe multi-agent interactions. We demonstrate that safety-critical control constraints can be effectively learned from data under relaxed assumptions regarding data relevance and observational uncertainty. Building on the learned safety representations, we design an impact-aware safe controller that improves both task efficiency and collective safety in urban and highway driving scenarios. Future directions include extending the method to systems with high relative degree and validating its performance through real-world experiments.

\vspace{-.5cm}

\nocite{*}  %
\bibliographystyle{IEEEtran}
\bibliography{cleanbib}

\end{document}